\documentclass[conference]{IEEEtran}
\IEEEoverridecommandlockouts

\usepackage{cite}
\usepackage{amsmath,amssymb,amsfonts}
\usepackage{graphicx}
\usepackage{array,tabularx,booktabs}
\usepackage{listings}
\usepackage{xcolor}
\usepackage{tikz}
\usepackage{pgfplots}
\pgfplotsset{compat=1.18}
\usepackage{textcomp}
\usepackage{url}
\usepackage{tikz}
\usetikzlibrary{calc}
\usepackage{hyperref}
\usepackage{fontawesome5}

\def\BibTeX{{\rm B\kern-.05em{\sc i\kern-.025em b}\kern-.08em
    T\kern-.1667em\lower.7ex\hbox{E}\kern-.125emX}}

\definecolor{promptbg}{RGB}{248,249,252}
\definecolor{promptborder}{RGB}{70,105,170}
\definecolor{cfpblue}{RGB}{45,105,180}
\definecolor{offgreen}{RGB}{55,145,90}
\definecolor{costred}{RGB}{190,70,70}
\definecolor{initmsg}{RGB}{232,240,254}
\definecolor{respmsg}{RGB}{235,247,237}
\definecolor{warnmsg}{RGB}{255,244,229}
\definecolor{awardmsg}{RGB}{242,242,242}
\lstdefinestyle{promptstyle}{
    basicstyle=\ttfamily\scriptsize,
    backgroundcolor=\color{promptbg},
    frame=single,
    rulecolor=\color{promptborder},
    framesep=2pt,
    breaklines=true,
    columns=fullflexible,
    keepspaces=true,
    showstringspaces=false,
    xleftmargin=1mm,
    xrightmargin=1mm,
    aboveskip=3pt,
    belowskip=3pt,
    captionpos=b
}

\newcommand{\papergraphic}[2][]{%
  \IfFileExists{#2}{\includegraphics[#1]{#2}}{%
    \fbox{\parbox[c][3.0cm][c]{0.94\linewidth}{\centering
    Figure available in the replication package.}}}}

\begin{document}

\title{Multi-Agent Scheduling with LLM-Assisted Contract Net Negotiation for Stream Processing in Mobile Edge Computing}

\author{
\IEEEauthorblockN{Sabeur Lajili}
\IEEEauthorblockA{\textit{University of Sousse, Tunisia}\\
sabeur.lajili@gmail.com}
\and
\IEEEauthorblockN{Zaki Brahmi}
\IEEEauthorblockA{\textit{University of Sousse, Tunisia}\\
zakibrahmi@gmail.com}
}

\maketitle

\begin{abstract}
Stream-processing systems increasingly operate across heterogeneous mobile edge--cloud infrastructures, where workload volatility, resource contention, and stringent quality-of-service (QoS) requirements complicate decentralized scheduling. This paper proposes \emph{MAS-DecStream}, whose main contribution is \emph{LLM-MR-CNP}: an extension of the classical Contract Net Protocol with semantic CFP formulation, progressive context disclosure, multi-round proposal revision, negotiation memory, and deterministic validation. Edge-cluster agents refine natural-language offloading proposals from local observations, predicted resource states, and qualitative runtime context, while hard resource and QoS constraints remain deterministic. Experiments derived from the Alibaba ASI Trace evaluate the extension at three levels: single- versus multi-round CNP, rule-based versus LLM-assisted refinement, and fixed-model single- versus multi-round negotiation. Under the evaluated configurations, MAS-DecStream reduces latency violations to 3\%, eliminates resource overcommitment, reaches a conflict-resolution rate of 0.91 with 20 agents, and improves utility by up to 22\% over the multi-round rule-based baseline. A separate 25-case evaluation shows model- and prompt-dependent accuracy--cost trade-offs. The results provide initial evidence that multi-round CNP refinement is the principal protocol-level gain, with LLM assistance adding value for qualitative and uncertain runtime context.
\end{abstract}


\section{Introduction}

Stream-processing systems support continuous analytics for latency-sensitive applications such as smart cities, intelligent transportation, and the Industrial Internet of Things. Scheduling these applications across heterogeneous edge--cloud infrastructures remains difficult because workloads evolve rapidly, resources are geographically distributed, and each edge cluster observes only part of the global state~\cite{lajili2025federated,wang2024deep}. Independent local decisions can therefore select the same destination, overload scarce resources, delay high-priority streams, and violate QoS constraints.

To address these challenges, recent studies have pushed scheduling decisions closer to the network edge using federated learning (FL)~\cite{lajili2025federated,zhao2024federated}, reinforcement learning (RL), and deep learning (DL)~\cite{cheng2023deep,jayanetti2022deep,wang2024deep} to predict workload variations and optimize distributed task placement. While these methods improve online adaptability, many still rely on centralized aggregation, introducing communication bottlenecks, single points of failure, and security risks~\cite{lajili2025federated}. Although FL alleviates this dependency, it often incurs significant communication and learning overhead while providing only partial visibility of the global system state~\cite{savazzi2020federated}. Consequently, edge clusters may make locally optimal yet globally conflicting scheduling decisions, leading to resource contention, inefficient task placement, delayed execution of high-priority tasks, and QoS violations in latency-sensitive stream applications~\cite{lajili2025federated}.

Classical multi-agent mechanisms, including auctions and CNP, provide explicit decentralized interaction~\cite{gerkey2002sold,zhang2019contractnet,smith1980contractnet}, yet their fixed message and bidding rules may be difficult to adapt when requirements include heterogeneous quantitative constraints and qualitative warnings. LLM-based agents offer a complementary capability: they can interpret semantically rich context, generate explanations, and revise proposals across interaction rounds~\cite{qian2025scaling,sapkota2026agents}.

\begin{figure}[t]
\centering
\papergraphic[width=\columnwidth]{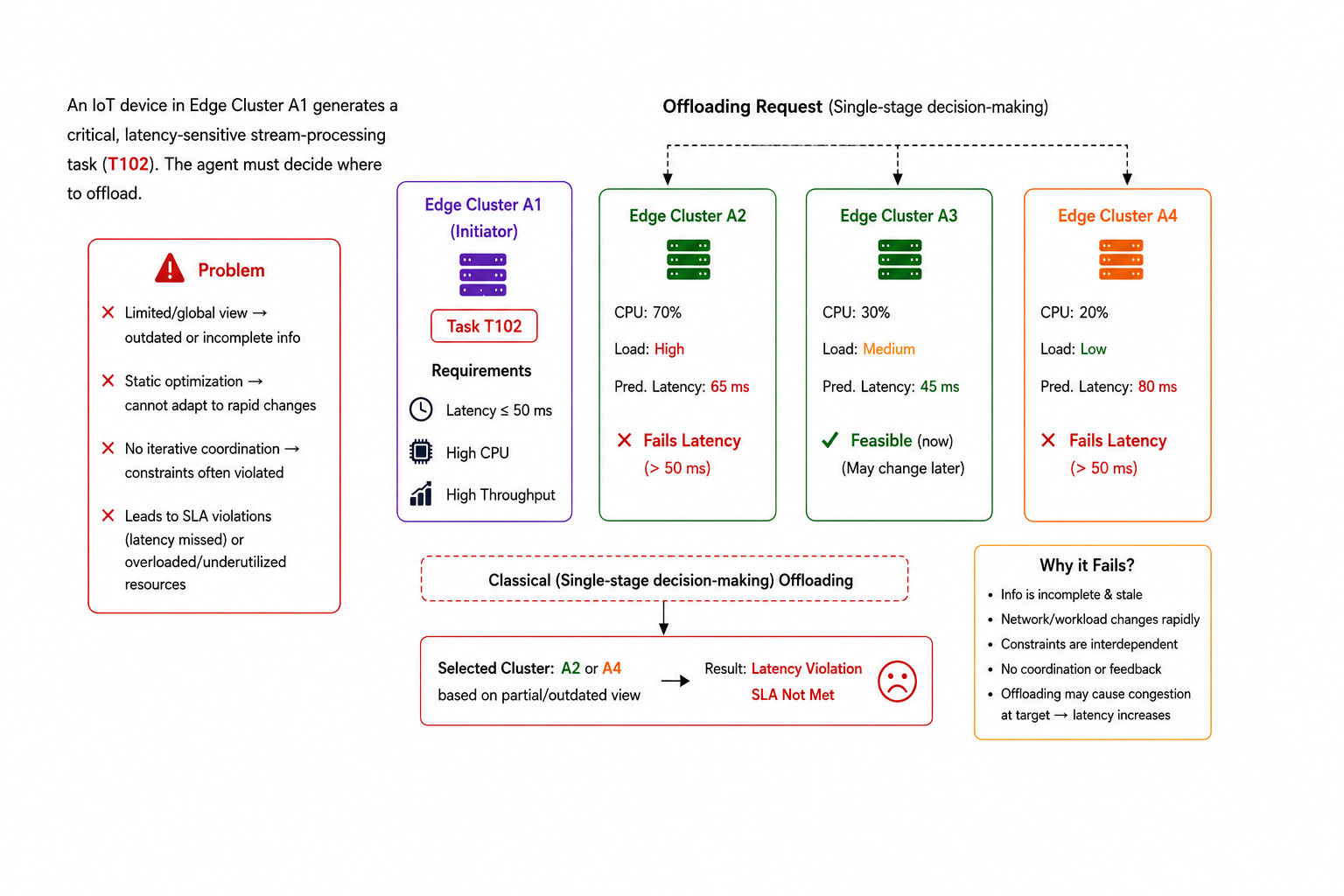}
\caption{Limitations of classical single-stage task-offloading decisions under partial system visibility.}
\label{fig:motivation}
\end{figure}

As illustrated in Figure~\ref{fig:motivation}, independently acting clusters may select the same apparently feasible destination, creating contention and QoS violations that cannot be anticipated from local observations alone. This motivates extending the classical single announcement--award CNP cycle with explicit inter-cluster proposal refinement before the final allocation.

This paper investigates this integration through \emph{MAS-DecStream}. Each representative edge-cluster agent observes its local cluster, uses tools for monitoring and validation, and negotiates task migration with neighboring agents. The proposed \emph{LLM-assisted Multi-Round CNP} (LLM-MR-CNP) progressively discloses additional workload and QoS context only when the first-round proposals do not yield a clear destination. The final allocation remains constrained by deterministic feasibility and utility checks.

The contributions are threefold:
\textbf{\faProjectDiagram~(1) LLM-MR-CNP}, an explicit extension of the classical CNP that introduces semantic CFP generation, progressive disclosure, iterative proposal revision, task-scoped negotiation memory, and bounded termination;
\textbf{\faRobot~(2) Hybrid agent architecture}, in which LLMs interpret contextual information and generate negotiation messages, while deterministic tools enforce resource, deadline, and utility constraints; and
\textbf{\faChartBar~(3) Comprehensive evaluation}, which distinguishes the effect of multi-round CNP refinement from the incremental contribution of LLM-assisted contextual reasoning, while jointly reporting scheduling quality and negotiation overhead.

In this paper, we address the following research questions:
\begin{itemize}
\item \textbf{RQ1:} How does extending single-round CNP with multi-round proposal refinement affect latency violations, utility, and coordination cost under workload drift?
\emph{Multi-round refinement reduces the latency-violation rate from 0.53 to 0.37, while LLM-assisted refinement further reduces it to 0.03 and increases utility from 1.03 to 1.61, at the cost of additional negotiation messages.}

\item \textbf{RQ2:} What incremental benefit does LLM-assisted contextual refinement provide over rule-based multi-round CNP under concurrent requests?  
\emph{Both multi-round approaches eliminate resource overcommitment, while LLM assistance increases conflict resolution from 0.86 to 0.91 with 20 agents and improves utility by up to 22\%.}

\item \textbf{RQ3:} How do negotiation depth, LLM choice, and prompting strategy affect CFP quality, offloading accuracy, latency, and token consumption?  
\emph{For a fixed large LLM, multi-round negotiation increases offloading accuracy from 0.71 to 0.88, while the best model--prompt configuration reaches 1.00 accuracy, although with higher latency and token consumption.}

\end{itemize}

The remainder of this paper is organized as follows. Section~\ref{sec:related} reviews related work on decentralized scheduling, LLM-assisted optimization, and multi-agent negotiation. Section~\ref{sec:model} presents the system model and scheduling objective. Section~\ref{sec:method} introduces the MAS-DecStream architecture and the proposed LLM-MR-CNP protocol. Section~\ref{sec:results} describes the experimental methodology, reports the results for the three research questions, and discusses the main findings and threats to validity. Finally, Section~\ref{sec:conclusion} concludes the paper and outlines future research directions.

\section{Related Work}\label{sec:related}
\subsection{Decentralized Scheduling and Task Offloading}
Edge--cloud scheduling has been studied through heuristic, optimization, reinforcement-learning, and federated-learning formulations. Recent DRL schedulers optimize latency, system load, energy, or execution cost under heterogeneous resource conditions~\cite{wang2024deep,cheng2023deep,jayanetti2022deep}, while federated approaches distribute model training and reduce direct sharing of operational data~\cite{lajili2025federated,zhao2024federated}. These methods improve adaptation, but coordination is commonly represented through a learned policy, a shared optimizer, or a fixed numerical exchange. Consequently, the scheduling logic may not explicitly expose how independently acting clusters reconcile simultaneous requests, revise offers after new information, or explain why a previously feasible destination becomes unsafe.

Classical multi-agent coordination provides explicit alternatives. Auctions support decentralized resource allocation through competing bids~\cite{gerkey2002sold}, whereas CNP decomposes task allocation into announcement, proposal, award, and rejection phases~\cite{smith1980contractnet,zhang2019contractnet}. Their message semantics and evaluation policies are transparent, but typically predefined. This becomes restrictive when a scheduling request combines numerical constraints with qualitative context, such as a reliability warning, a privacy condition, or an uncertain workload forecast. MAS-DecStream therefore retains CNP as the interaction backbone while extending proposal interpretation and refinement beyond fixed numerical bids.

\subsection{LLM- and Agentic-AI-Based Scheduling}
LLMs have recently been integrated into planning, optimization, and resource-allocation workflows. Mongaillard et al.~\cite{mongaillard2024large} use LLM-assisted agents to translate user requirements into electric-vehicle charging decisions, and Zhang et al.~\cite{zhang2026agentic} introduce an agentic framework for UAV-assisted logistics scheduling. In MEC, COMLLM formulates offloading as language-conditioned sequential decision making~\cite{yang2026multi}; Ma et al.~\cite{ma2025multi} study multi-tier deployment of LLM inference across heterogeneous edge--cloud resources; and AWTO optimizes placement of LLM-driven agentic workflows under latency constraints~\cite{yu2026awto}. Wang et al.~\cite{wang2025maef} further show that multiple LLM agents can collaboratively generate, evaluate, and refine candidate schedules.

These studies establish that LLMs can interpret high-level requirements and support optimization, but their primary focus is usually user-request translation, workflow placement, inference deployment, or centralized schedule search. They do not directly study an explicit decentralized contract protocol in which autonomous edge-cluster representatives exchange offers, revise them under newly disclosed context, and preserve hard resource and QoS guarantees through deterministic validation.

\subsection{LLM-Based Multi-Agent Cooperation and Negotiation}
Agentic-AI research has progressed from prompt-driven reasoning to stateful, role-based orchestration. ReAct combines reasoning with actions~\cite{yao2023react}; Generative Agents and CAMEL demonstrate memory-supported and role-playing interaction~\cite{park2023generative,li2024camel}; and AutoGen and MetaGPT coordinate specialized agents through structured conversation and workflows~\cite{wu2023autogen,hong2024metagpt}. More recent systems employ deliberation or negotiation for consensus and conflict resolution. Multi-Agent Debate improves reasoning through iterative critique~\cite{du2023debate}, CoLMDriver applies LLM negotiation to cooperative driving~\cite{liu2025colmdriver}, and TeamFusion supports open-ended teamwork among heterogeneous agents~\cite{liu2026teamfusion}. Scaling studies also show that communication structure and role assignment strongly affect collective performance~\cite{qian2025scaling}.

Table~\ref{tab:positioning} summarizes the closest research directions. MAS-DecStream is positioned at their intersection. It targets resource-constrained stream migration, uses an explicit decentralized negotiation protocol, supports multi-round contextual refinement, and separates LLM-generated semantic decisions from deterministic feasibility and utility checks. Its novelty is therefore integrative rather than the invention of CNP or LLM agents in isolation.

\begin{table}[h]
\centering
\caption{Positioning relative to representative research directions.}
\label{tab:positioning}
\scriptsize
\setlength{\tabcolsep}{2.5pt}
\begin{tabularx}{\columnwidth}{p{2cm}p{1.55cm}X}
\toprule
Direction & Coordination & Main distinction from MAS-DecStream \\
\midrule
FL/DRL edge scheduling~\cite{lajili2025federated,wang2024deep} & Learned policy & No explicit proposal revision among cluster representatives \\
Language-conditioned MEC offloading~\cite{yang2026multi} & Sequential reasoning & Offloading decisions are not organized as multi-agent CNP negotiation \\
Agentic workflow placement~\cite{yu2026awto} & Optimized placement & Focuses on agent-workflow execution rather than stream migration negotiation \\
LLM multi-agent optimization~\cite{wang2025maef} & Iterative search & Refines schedules, but not through hard-validated edge-cluster contracts \\
LLM negotiation systems~\cite{liu2025colmdriver,liu2026teamfusion} & Natural-language interaction & Do not target heterogeneous MEC stream resources and QoS constraints \\
\textbf{MAS-DecStream} & \textbf{LLM-MR-CNP} & \textbf{Contextual refinement with deterministic scheduling safeguards} \\
\bottomrule
\end{tabularx}
\end{table}

\section{System Model and Objective}\label{sec:model}
The edge--cloud environment is modeled as a communication graph $\mathcal{G}=(\mathcal{A},\mathcal{E})$, where $\mathcal{A}=\{A_1,\ldots,A_N\}$ is the set of representative edge-cluster agents and $\mathcal{E}$ contains their communication links. Agent $A_i$ observes only its represented cluster, including current and predicted CPU, memory, bandwidth, latency, workload, and energy conditions. A stream task $T_k\in\mathcal{T}$ specifies resource demands, a latency deadline, importance, and optional compatibility or privacy requirements.

Let $D_k=A_j$ denote assigning $T_k$ to the cluster represented by $A_j$, and let $M^r$ denote the messages exchanged through $r$ negotiation rounds. The scheduling  objective is (eq. \ref{eq:1})
\begin{equation} \label{eq:1}
(D^*,r^*)=\arg\min_{D,r} OF(D,M^r),
\end{equation}
where
\begin{equation*}
\begin{aligned}
OF(D,M^r)=&\;\alpha_1\widetilde{Latency}(D)
+\alpha_2\widetilde{Energy}(D)\\
&+\alpha_3\bigl(1-LBD_{\mathrm{total}}(D)\bigr)
+\alpha_4\widetilde{CO}(M^r)
\end{aligned}
\end{equation*}

The weights satisfy $\alpha_m\geq0$ and
$\sum_m\alpha_m=1$. $LBD_{\mathrm{total}}(D)\in[0,1]$ denotes the overall load-balancing degree induced by assignment $D$, where values
closer to one indicate a more balanced workload distribution;
thus, $1-LBD_{\mathrm{total}}(D)$ represents the load-imbalance
penalty. The term $\mathrm{CO}(M^r)$ denotes the coordination
overhead of the negotiation, including the number and size of
the exchanged messages and, when applicable, the latency of the
negotiation rounds. A destination is feasible only if its
post-allocation CPU, memory, and bandwidth remain within
capacity and the task's deadline, compatibility, privacy, and
execution requirements are satisfied. 

The objective function is evaluated by the agent's deterministic reasoning tools to rank proposed bids. Furthermore, it uses LLM to interpret contextual information and generate or refine the CNP messages used to obtain the required inputs. The objective therefore captures the trade-off between placement quality, load balance, and the
communication overhead introduced by proposal-refinement rounds.

\section{MAS-DecStream}\label{sec:method}
\subsection{Hybrid Agent Architecture}
\begin{figure}[h]
\centering
\papergraphic[width=\columnwidth]{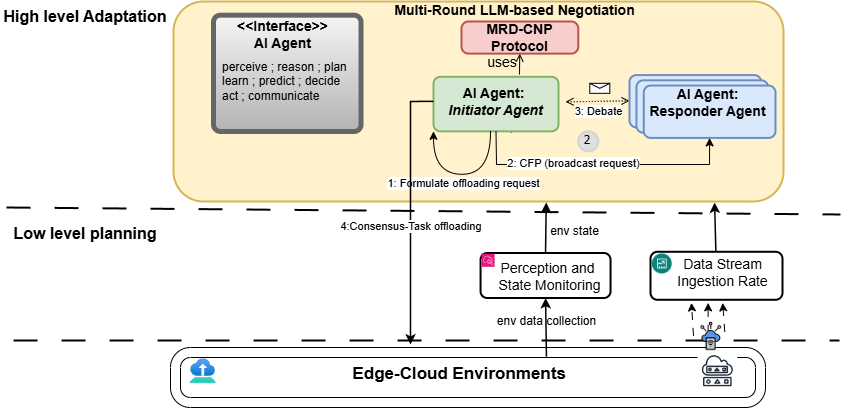}
\caption{MAS-DecStream separates local monitoring and execution from collaborative, LLM-assisted decision making.}
\label{fig:arch}
\end{figure}

Figure~\ref{fig:arch} presents two interacting layers. The execution layer monitors runtime telemetry, predicts near-future load, and applies migration decisions. The collaborative layer contains stateful edge-cluster agents orchestrated using LangGraph~\cite{langgraph2024}. Each agent combines: (i) a local observation and predicted state; (ii) task and QoS knowledge; (iii) task-scoped negotiation memory; (iv) an LLM for contextual interpretation and message generation; and (v) tools for state retrieval, requirement validation, and utility computation. The LLM may interpret warnings and refine a proposal, but measurements, hard constraints, and numerical ranking remain deterministic.

\subsection{LLM-Assisted Multi-Round CNP}
The central contribution is a protocol extension rather than unconstrained agent conversation. As summarized in Table~\ref{tab:cnp_extension}, \textbf{LLM-MR-CNP} preserves the classical roles of initiator, responder, CFP, proposal, and award, but changes how information is represented, revised, validated, and terminated. LLM-MR-CNP extends classical CNP~\cite{smith1980contractnet} through five steps. First, an overloaded initiator broadcasts a compact CFP containing the essential task demand and scheduling intent. Second, each responder inspects its current and predicted state and returns a proposal or refusal. Third, the initiator discards responses that violate hard constraints. Fourth, when several feasible candidates remain close or a proposal is uncertain, the initiator progressively discloses additional context, such as forecast load, task priority, or a reliability warning, and requests revised proposals only from the remaining candidates. Finally, negotiation terminates when one feasible candidate remains, a stable best candidate emerges, or the maximum number of rounds is reached. The selected destination is
\begin{equation}\label{eq:selection}
A^*=\arg\max_{A_j\in\mathcal{A}_{\mathrm{feasible}}(T_k)} U_j(T_k).
\end{equation}
If no feasible proposal exists, the task is deferred or forwarded to the cloud. Listing~\ref{lst:prompt} summarizes the two role templates. They constrain the agents to retrieved telemetry and request concise, machine-checkable CNP actions rather than unrestricted dialogue.

\begin{lstlisting}[style=promptstyle,caption={Compact initiator and responder templates.},label={lst:prompt}]
INITIATOR
Role: edge-cluster agent under resource or QoS pressure.
1. Retrieve the current local state.
2. If local execution is unsafe, generate a concise CFP.
3. Include only observed task and telemetry information.
Output: task intent, demand, deadline, and requested checks.

RESPONDER
Role: candidate edge-cluster agent evaluating the CFP.
1. Retrieve current and predicted local capacity.
2. Check feasibility and latency risk with local tools.
3. Return [propose] or [refuse] and the binding reason.
Output: decision, validated capacity, and short rationale.
\end{lstlisting}

\begin{figure}[h]
\centering
\begin{tikzpicture}[
  msg/.style={rounded corners=7pt, align=left, font=\scriptsize\sffamily,
              inner xsep=6pt, inner ysep=4pt, text width=0.58\columnwidth},
  msgL/.style={msg, fill=initmsg,  draw=cfpblue!35},
  msgW/.style={msg, fill=warnmsg,  draw=orange!60!black!35},
  msgR/.style={msg, fill=respmsg,  draw=offgreen!55!black!30},
  sys/.style={rounded corners=8pt, align=center, font=\scriptsize\sffamily,
              inner xsep=8pt, inner ysep=4pt, fill=awardmsg, draw=black!25,
              text width=0.72\columnwidth},
  avL/.style={circle, fill=cfpblue, text=white, font=\tiny\sffamily\bfseries,
              inner sep=0pt, minimum size=11.5pt},
  avR/.style={avL, fill=offgreen!60!black},
]
\def\vgap{5pt}   
\def\agap{4pt}   
\coordinate (L) at (0,0);
\coordinate (R) at (0.96\columnwidth,0);
\newcommand{\hdr}[2]{{\tiny\bfseries\color{#1}#2}\\[0.5pt]}

\node[msgL, anchor=north west] (m1) at ($(L)+(16pt,0)$)
  {\hdr{cfpblue!80!black}{A$_0$ \,\textperiodcentered\, cfp}%
   I need to migrate a high-priority ECG stream and its latency deadline must be preserved. Can you host this task?};
\node[avL, anchor=north east] at ($(m1.north west)+(-\agap,0)$) {A$_0$};

\node[msgR, anchor=north east] (m2) at ($(m1.south -| R)+(-16pt,-\vgap)$)
  {\hdr{offgreen!45!black}{A$_1$ \,\textperiodcentered\, propose}%
   Yes --- my resources are sufficient and I expect my load to stay
   stable, so I can host the stream without degrading its QoS.};
\node[avR, anchor=north west] at ($(m2.north east)+(\agap,0)$) {A$_1$};

\node[msgR, anchor=north east] (m3) at ($(m2.south -| R)+(-16pt,-\vgap)$)
  {\hdr{offgreen!45!black}{A$_2$ \,\textperiodcentered\, propose}%
   I can host it too: my predicted latency is well below the deadline. However, my near-future workload is uncertain, so treat this as a provisional proposal.};
\node[avR, anchor=north west] at ($(m3.north east)+(\agap,0)$) {A$_2$};

\node[msgW, anchor=north west] (m4) at ($(m3.south -| L)+(16pt,-\vgap)$)
  {\hdr{orange!60!black}{A$_0$ \,\textperiodcentered\, refine}%
   $A_2$, your proposal currently ranks best, but my reliability monitor
   just reported a traffic spike building near you. Please reassess your
   predicted latency under this warning.};
\node[avL, anchor=north east] at ($(m4.north west)+(-\agap,0)$) {A$_0$};

\node[msgR, anchor=north east] (m5) at ($(m4.south -| R)+(-16pt,-\vgap)$)
  {\hdr{offgreen!45!black}{A$_2$ \,\textperiodcentered\, refuse}%
   You're right --- factoring in that spike, my predicted latency may
   exceed the deadline. I withdraw my proposal.};
\node[avR, anchor=north west] at ($(m5.north east)+(\agap,0)$) {A$_2$};

\node[sys, anchor=north] (m6) at ($(m5.south -| 0.5\columnwidth,0)+(0,-\vgap)$)
  {\textbf{Award} \,\textperiodcentered\, $A_2$'s withdrawal removes the
   initially best bid; deterministic validation confirms $A_1$'s proposal
   meets the deadline, and the ECG stream is awarded to $A_1$.};
\end{tikzpicture}
\caption{ Illustration of LLM-MR-CNP for an ECG-stream migration. Agents exchange natural-language messages tagged with CNP performatives
(\textsf{cfp}, \textsf{propose}, \textsf{refine}, \textsf{refuse}) rather than fixed-format bids. $A_2$ initially submits the best proposal. Classical single-round CNP would have awarded the stream to $A_2$, whereas LLM-MR-CNP redirects it to the feasible node $A_1$.}
\label{fig:protocol_chat}

\end{figure}
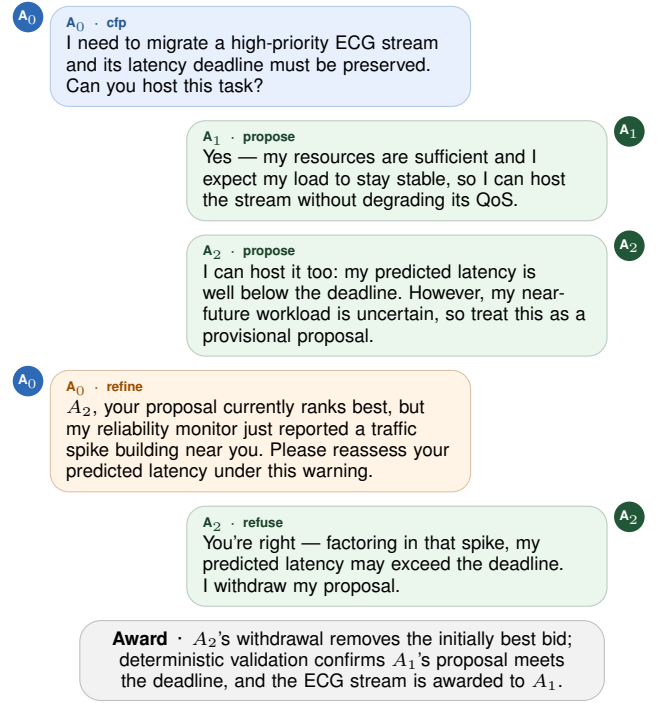

Figure~\ref{fig:protocol_chat} illustrates how these mechanisms alter a concrete CNP exchange. Two responders initially consider an ECG stream feasible, but one proposal is provisional because its forecast is uncertain. Classical CNP would normally proceed to the award after this first collection phase. LLM-MR-CNP instead discloses only the missing reliability context, requests a targeted revision, and then applies deterministic validation before the award. This is negotiation rather than peer-to-peer debate: responders revise their own bids, while the initiator controls disclosure, validation, and termination.

Table~\ref{tab:cnp_extension} makes the extension explicit. Classical CNP normally closes the allocation after one proposal-collection phase; in LLM-MR-CNP, a responder may issue a provisional proposal and later revise or withdraw it after the initiator discloses additional forecast, priority, or reliability context. The final award is nevertheless based on validated feasibility and objective utility rather than rhetorical persuasiveness or LLM confidence.

\begin{table}[h]
\centering
\caption{Classical CNP versus the proposed LLM-MR-CNP extension.}
\label{tab:cnp_extension}
\scriptsize
\setlength{\tabcolsep}{2pt}
\begin{tabularx}{\columnwidth}{p{1.20cm}p{2.20cm}X}
\toprule
Aspect & Classical CNP & LLM-MR-CNP extension \\
\midrule
CFP & Fixed task announcement & Semantic CFP from task, QoS, and observed context \\
Interaction & One proposal--award cycle & Bounded proposal-refinement rounds \\
Information & Predefined fields disclosed once & Progressive disclosure to unresolved candidates \\
Proposal & Bid/refusal is normally final & Proposal may be provisional, revised, or withdrawn \\
Memory & Current exchange & Task-scoped history across rounds \\
Safety & Fixed bid-evaluation rule & LLM interpretation followed by hard validation \\
Termination & After collecting proposals & Clear best candidate, stable decision, or round limit \\
\bottomrule
\end{tabularx}
\end{table}

\paragraph{Protocol safeguards.}
Each message is task-scoped and recorded in the negotiation history. Malformed responses, unsupported numerical claims, or proposals that contradict tool outputs are rejected or replaced by a deterministic fallback. A round terminates when one feasible candidate remains, the leading candidate is sufficiently separated from alternatives, the decision stabilizes across refinements, or the maximum round budget is reached. These safeguards limit the effect of stochastic generation while preserving the LLM's role in interpreting qualitative context and producing concise coordination messages.

\section{Experimental Evaluation}\label{sec:results}
\subsection{Experimental Methodology}
The experiments are organized around the protocol extension. RB-SR-CNP versus RB-MR-CNP evaluates the move from the classical single-round cycle to iterative refinement. RB-MR-CNP versus MAS-DecStream evaluates the complete semantic-context and LLM-assisted layer. Under a fixed large LLM, single- versus multi-round negotiation further tests whether the interaction structure itself improves the final award. Prompting experiments then examine how reliably the semantic CFP and responder components can be instantiated.

\subsubsection{Baselines and Controlled Comparison}
We compare three scheduling pipelines. \textbf{RB-SR-CNP} uses a classical one-round announcement--proposal--award exchange. \textbf{RB-MR-CNP} introduces iterative proposal refinement and updated quantitative forecasts but does not use an LLM. \textbf{MAS-DecStream} retains the multi-round workflow and adds LLM-assisted CFP interpretation, qualitative-context handling, and natural-language proposal revision. All conditions receive the same task instances, candidate clusters, current resource states, hard feasibility rules, and deterministic utility function. This design controls the scheduling environment, although MAS-DecStream still differs from RB-MR-CNP in both access to qualitative context and its interpretation mechanism; Scenarios~1 and~2 are therefore pipeline comparisons rather than component-level causal ablations.

\subsubsection{Trace-Derived Workloads and Reference Labels}
The workload is derived from the Alibaba ASI Trace 2026 job-execution summary~\cite{asi_trace_2026}, which contains large-scale production AI workload observations. We sample 1,000 records and enrich them with stream- and MEC-specific attributes required by the scheduling problem, including CPU, memory, bandwidth, latency deadline, priority, and contextual requirements. The resulting benchmark contains 1,000 workload records and five heterogeneous candidate-cluster snapshots per workload, yielding 5,000 task--cluster combinations. The five profiles represent high-compute, energy-efficient, low-latency, privacy-enabled, and overloaded-source conditions (Table~\ref{tab:profiles}).

Reference labels specify the expected CFP intent, acceptable responder actions, candidate-refinement focus, and final destination used in Scenario~3. Label construction follows the benchmark's hard feasibility and objective criteria, and labels are withheld from the LLMs during inference. The prompt templates receive task and cluster descriptions only. Generic demonstrations used by few-shot prompts are separated from the evaluated cases. The generated workloads, labels, prompts, and raw outputs are included in the replication package.\footnote{\url{https://github.com/MythesisProject2024/MAS_DecStream}}

\subsubsection{Implementation and Evaluation Measures}
Experiments are implemented in Python using LangGraph~\cite{langgraph2024} and Ollama. The scheduling scenarios use gpt-4o-mini for the LLM-assisted pipeline, while the prompting study evaluates local Llama3 and cloud-routed GPT-OSS:20B, GLM-5.2, DeepSeek-V4-Pro, and Gemini-3-Flash configurations. Single-round negotiation is limited to one exchange; the proposed workflow permits up to three rounds and progressively narrows the responder set. Exact prompt templates and model identifiers are released with the benchmark.

We report complementary system and reasoning measures. The \emph{latency-violation rate} is the fraction of scheduled windows or tasks exceeding their deadline. The \emph{overcommitment rate} measures the fraction of clusters whose cumulative awarded demand exceeds capacity, whereas the \emph{conflict-resolution rate} captures the share of concurrent allocation conflicts resolved without infeasible assignment. \emph{Global utility} aggregates valid allocations and QoS outcomes according to the common objective, and \emph{collaboration cost} counts negotiation messages. Scenario~3 additionally evaluates CFP intent and required-context coverage, responder action accuracy, final-host accuracy, end-to-end decision time, and estimated prompt and completion tokens. CFP and responder measures diagnose intermediate reasoning, while final-host accuracy and QoS outcomes remain the primary end-to-end indicators.

\begin{table}[h]
\centering
\caption{Representative cluster profiles.}
\label{tab:profiles}
\scriptsize
\setlength{\tabcolsep}{3pt}
\begin{tabular}{cll}
\toprule
Agent & Cluster role & Representative suitability \\
\midrule
$A_0$ & Offloading initiator & Locally overloaded source \\
$A_1$ & High-compute & Compute-intensive streams \\
$A_2$ & Energy-efficient/trusted & Energy-aware analytics \\
$A_3$ & Stable low-latency & ECG and AR streams \\
$A_4$ & Privacy-enabled & Sensitive video analytics \\
\bottomrule
\end{tabular}
\end{table}

\subsection{RQ1: Migration Under Data Drift}
An unmodelled workload spike creates overload and deadline risk over a 300-s horizon sampled every 10~s. The initiator triggers migration in 23 of the 30 windows. MAS-DecStream uses gpt-4o-mini to interpret the proposals and qualitative drift context before deterministic validation.

\begin{table}[h]
\centering
\caption{Migration under data drift.}
\label{tab:drift}
\scriptsize
\setlength{\tabcolsep}{2.5pt}
\begin{tabular}{lccccc}
\toprule
Method & Trig. & Lat. viol. & Utility & Cost & LLM time \\
\midrule
RB-SR-CNP & 23 & 0.53 & 1.11 & 92 & -- \\
RB-MR-CNP & 23 & 0.37 & 1.03 & 184 & -- \\
MAS-DecStream & 23 & 0.03 & 1.61 & 276 & 83.56~s \\
\bottomrule
\end{tabular}
\end{table}

Table~\ref{tab:drift} provides the first validation of the CNP extension. RB-SR-CNP has the lowest coordination cost, but its one-shot bids are fixed by the state available at announcement time and produce violations in more than half of the evaluated windows. Replacing this single proposal--award cycle with rule-based refinement reduces the violation rate from 0.53 to 0.37, confirming that iterative proposal revision is useful even without an LLM. MAS-DecStream further reduces the rate to 0.03 and obtains the highest utility by allowing the refinement message to incorporate qualitative drift context before deterministic validation.

The improvement is not free: MAS-DecStream uses three times the single-round collaboration cost and introduces 83.56~s of reported LLM reasoning time over the scenario. This overhead is important for latency-sensitive systems and motivates selective invocation rather than using an LLM at every monitoring window. In particular, the execution layer can handle routine decisions deterministically and activate contextual negotiation only when prediction uncertainty, conflicting proposals, or non-numerical warnings make the ordinary policy insufficient. Because MAS-DecStream differs from RB-MR-CNP in both qualitative context and LLM assistance, the result supports the complete context-aware pipeline; it does not, by itself, quantify the causal contribution of the LLM.

\IfFileExists{fig/sensitivity_delta_utility.csv}{%
\subsubsection{Sensitivity to Drift and Forecast Uncertainty}
To characterize the operating conditions under which semantic contextual
refinement provides value beyond rule-based multi-round CNP, we vary
unmodelled workload drift and forecast uncertainty. Drift severity
$d\in\{0,10,20,30,40,50\}\%$ represents additional realized resource demand
not captured by the prediction model. Forecast uncertainty
$e\in\{0,10,20,30,40\}\%$ is introduced by perturbing the predicted CPU,
memory, bandwidth, and latency values before negotiation. For every $(d,e)$
configuration, RB-MR-CNP and MAS-DecStream are evaluated on the same
repeated seeded workload trajectories. We report
$\Delta U=\overline{U}_{\mathrm{MAS}}-\overline{U}_{\mathrm{RB\text{-}MR}}$;
positive values indicate an advantage for LLM-assisted contextual refinement.

\begin{figure}[t]
\centering
\begin{tikzpicture}
\begin{axis}[
    width=0.78\columnwidth,
    height=3.8cm,
    xlabel={Forecast uncertainty (\%)},
    ylabel={Unmodelled drift (\%)},
    xmin=-5, xmax=45,
    ymin=-5, ymax=55,
    xtick={0,10,20,30,40},
    ytick={0,10,20,30,40,50},
    tick label style={font=\scriptsize},
    label style={font=\scriptsize},
    enlargelimits=false,
    axis on top,
    colorbar,
    colorbar style={
        ylabel={Mean utility difference $\Delta U$},
        ylabel style={font=\scriptsize},
        tick label style={font=\scriptsize}
    },
    nodes near coords*={
        \pgfmathprintnumber[fixed,precision=2]{\pgfplotspointmeta}
    },
    every node near coord/.append style={font=\tiny}
]
\addplot[
    matrix plot*,
    mesh/cols=5,
    point meta=explicit
]
table[
    x=uncertainty,
    y=drift,
    meta=deltaU,
    col sep=comma
] {fig/sensitivity_delta_utility.csv};
\end{axis}
\end{tikzpicture}
\caption{Sensitivity of the proposed CNP extension to workload drift and
forecast uncertainty. Each cell reports the mean utility difference between
MAS-DecStream and RB-MR-CNP over identical seeded workload trajectories.
Positive values indicate an advantage for LLM-assisted contextual refinement.}
\label{fig:sensitivity_heatmap}
\end{figure}

Figure~\ref{fig:sensitivity_heatmap} provides an operating-region view of
the incremental value of semantic contextual refinement. Positive cells show
where MAS-DecStream improves utility over RB-MR-CNP, values near zero indicate
that numerical multi-round refinement is sufficient, and negative cells show
where the additional LLM layer does not compensate for its decisions. The grid
therefore identifies the drift--uncertainty regions in which semantic proposal
revision is most likely to justify its additional inference cost.
}{}

\subsection{RQ2: Concurrent Conflict Resolution}
We increase the setting from five agents and three requests to 20 agents and 12 concurrent requests. The tasks include ECG monitoring, video surveillance, and emergency alarms, creating contention between latency and priority requirements.

\begin{table}[h]
\centering
\caption{Conflict resolution under concurrent requests.}
\label{tab:conflict}
\scriptsize
\resizebox{\columnwidth}{!}{%
\begin{tabular}{cclccrr}
\toprule
Agents & Req. & Method & Conflict res. & Overcommit. & Utility & Cost/time (ms) \\
\midrule
5 & 3 & RB-SR-CNP & 0.00 & 0.20 & -6.00 & 12/160 \\
5 & 3 & RB-MR-CNP & 1.00 & 0.00 & 4.51 & 24/320 \\
5 & 3 & MAS-DecStream & 1.00 & 0.00 & 5.50 & 24/320 \\
\midrule
10 & 6 & RB-SR-CNP & 0.00 & 0.10 & -12.00 & 54/370 \\
10 & 6 & RB-MR-CNP & 0.90 & 0.00 & 10.77 & 108/700 \\
10 & 6 & MAS-DecStream & 0.95 & 0.00 & 12.08 & 108/840 \\
\midrule
20 & 12 & RB-SR-CNP & 0.00 & 0.05 & -23.00 & 216/850 \\
20 & 12 & RB-MR-CNP & 0.86 & 0.00 & 23.29 & 430/1600 \\
20 & 12 & MAS-DecStream & 0.91 & 0.00 & 24.24 & 433/1950 \\
\bottomrule
\end{tabular}}
\end{table}

The dominant result in Table~\ref{tab:conflict} is the value of the multi-round CNP extension. RB-SR-CNP makes awards without sufficient reconciliation of simultaneous demand, leading to persistent overcommitment and negative utility. Both multi-round methods reserve capacity across successive awards and eliminate overcommitment in every setting. Thus, successive validation, reservation, and proposal refinement constitute the principal source of robustness under concurrency.

MAS-DecStream adds smaller but consistent gains over RB-MR-CNP. Utility improves by approximately 22\%, 12\%, and 4\% for the 5-, 10-, and 20-agent settings, respectively, while conflict resolution increases by five percentage points at 10 and 20 agents. The diminishing relative gain suggests that, as the candidate set grows, deterministic feasibility and reservation already resolve most conflicts, leaving fewer decisions in which semantic interpretation can alter the outcome. At the same time, decision time increases from 700 to 840~ms at 10 agents and from 1600 to 1950~ms at 20 agents. Thus, LLM assistance is most defensible for ambiguous, high-priority, or context-dependent requests, whereas routine contention can remain under rule-based multi-round coordination.

\subsection{RQ3: LLM and Prompting Effects}
\subsubsection{Model Capability and Negotiation Depth}
Table~\ref{tab:ablation} compares a small local LLM, a larger LLM under single- and multi-round negotiation, and RB-MR-CNP over 25 cases.

\begin{table}[h]
\centering
\caption{Model and negotiation-depth comparison over 25 cases.}
\label{tab:ablation}
\scriptsize
\resizebox{\columnwidth}{!}{%
\begin{tabular}{lccccc}
\toprule
Configuration & CFP acc. & Offload acc. & Resp. acc. & Time (s) & Tokens \\
\midrule
Small LLM, multi-round & 0.46 & 0.78 & 0.69 & 186.18 & 48,927 \\
Large LLM, single-round & 0.56 & 0.71 & 0.60 & 20.10 & 19,100 \\
Large LLM, multi-round & 0.56 & 0.88 & 0.63 & 75.15 & 53,452 \\
RB-MR-CNP & -- & 0.82 & 0.64 & 65.01 & -- \\
\bottomrule
\end{tabular}}
\end{table}

The larger model improves CFP accuracy from 0.46 to 0.56 relative to the small model, but model size alone does not guarantee a better allocation: the large single-round configuration reaches only 0.71 offloading accuracy. Allowing refinement raises this value to 0.88, indicating that the interaction structure is at least as important as raw model capability. The improvement comes with a substantial cost: token consumption grows from 19,100 to 53,452 and decision time rises from 20.10 to 75.15~s. The large multi-round model is six points above RB-MR-CNP in final-host accuracy, while responder accuracy remains nearly unchanged. This pattern suggests that the main benefit is not uniformly better local accept/refuse judgments, but the initiator's ability to combine, refine, and validate multiple imperfect responses before the final award.

\subsubsection{Prompting Across LLMs}
We compare zero-shot, few-shot, CoT, ReAct, and hybrid prompts across five
LLMs. Table~\ref{tab:scenario4b_llm_comparison} reports all 30 evaluated
model--prompt configurations rather than only the best result for each model,
thereby exposing the variability of CFP formulation, responder decisions,
final offloading accuracy, inference time, and token consumption.

\begin{table}[h]
\centering
\caption{End-to-end MAS-DecStream negotiation comparison over 25 cases.}
\label{tab:scenario4b_llm_comparison}
\scriptsize
\setlength{\tabcolsep}{2.0pt}
\renewcommand{\arraystretch}{0.90}
\resizebox{\columnwidth}{!}{%
\begin{tabular}{llccccc}
\toprule
LLM & Prompting & CFP acc. & Offload acc. & Responder acc. & Time (s) & Tokens \\
\midrule
GPT-OSS:20B & Zero-shot & 0.56 & 0.84 & 0.63 & 75.15 & 53,452 \\
GPT-OSS:20B & Few-shot & 0.64 & 0.88 & 0.59 & 41.95 & 53,961 \\
GPT-OSS:20B & CoT & 0.75 & 0.84 & 0.66 & 63.20 & 55,531 \\
GPT-OSS:20B & CoT+few-shot & 0.76 & 0.92 & 0.57 & 64.63 & 51,528 \\
GPT-OSS:20B & ReAct & 0.71 & 0.92 & 0.63 & 44.99 & 60,829 \\
GPT-OSS:20B & ReAct+few-shot+CoT & 0.76 & 0.92 & 0.63 & 59.70 & 60,473 \\
\midrule
DeepSeek-V4-Pro & Zero-shot & 0.44 & 0.80 & 0.55 & 67.29 & 45,417 \\
DeepSeek-V4-Pro & Few-shot & 0.84 & 0.84 & 0.59 & 53.35 & 52,247 \\
DeepSeek-V4-Pro & CoT & 0.64 & 0.80 & 0.54 & 58.11 & 47,947 \\
DeepSeek-V4-Pro & CoT+few-shot & 0.72 & 0.84 & 0.56 & 54.82 & 45,035 \\
DeepSeek-V4-Pro & ReAct & 0.41 & 0.72 & 0.57 & 64.73 & 53,871 \\
DeepSeek-V4-Pro & ReAct+few-shot+CoT & 0.64 & 0.68 & 0.58 & 74.63 & 48,919 \\
\midrule
GLM-5.2 & Zero-shot & 0.51 & 0.76 & 0.48 & 56.41 & 60,649 \\
GLM-5.2 & Few-shot & 0.60 & 0.83 & 0.58 & 57.73 & 61,316 \\
GLM-5.2 & CoT & 0.69 & 0.80 & 0.59 & 68.62 & 59,809 \\
GLM-5.2 & CoT+few-shot & 0.52 & 0.83 & 0.60 & 90.56 & 52,131 \\
GLM-5.2 & ReAct & 0.74 & 0.92 & 0.56 & 102.50 & 67,184 \\
GLM-5.2 & ReAct+few-shot+CoT & 0.84 & 0.72 & 0.50 & 95.71 & 62,008 \\
\midrule
Gemini-3-Flash & Zero-shot & 0.44 & 0.82 & 0.68 & 59.14 & 55,494 \\
Gemini-3-Flash & Few-shot & 0.64 & 0.84 & 0.61 & 62.30 & 59,367 \\
Gemini-3-Flash & CoT & 0.52 & 0.92 & 0.67 & 62.52 & 58,296 \\
Gemini-3-Flash & CoT+few-shot & 0.54 & 0.92 & 0.63 & 64.99 & 53,276 \\
Gemini-3-Flash & ReAct & 0.43 & 0.92 & 0.65 & 60.83 & 56,733 \\
Gemini-3-Flash & ReAct+few-shot+CoT & 0.84 & 1.00 & 0.67 & 69.59 & 45,621 \\
\midrule
Llama3 & Zero-shot & 0.50 & 0.80 & 0.59 & 186.18 & 48,927 \\
Llama3 & Few-shot & 0.66 & 0.85 & 0.59 & 156.18 & 48,200 \\
Llama3 & CoT & 0.53 & 0.92 & 0.57 & 136.27 & 47,271 \\
Llama3 & CoT+few-shot & 0.67 & 0.92 & 0.60 & 120.20 & 44,300 \\
Llama3 & ReAct & 0.53 & 0.86 & 0.60 & 110.00 & 63,300 \\
Llama3 & ReAct+few-shot+CoT & 0.64 & 0.90 & 0.59 & 135.56 & 58,202 \\
\bottomrule
\end{tabular}}
\end{table}

Table~\ref{tab:scenario4b_llm_comparison} shows that the best prompting
strategy is model-dependent. Few-shot and hybrid prompts frequently improve
CFP coverage, but a better CFP does not necessarily produce a better final
destination. For example, ReAct reaches 0.92 offloading accuracy for
GPT-OSS:20B and GLM-5.2, whereas adding few-shot and CoT to ReAct reduces
DeepSeek-V4-Pro from 0.72 to 0.68 and GLM-5.2 from 0.92 to 0.72 despite the
higher CFP accuracy of the latter configuration. Gemini-3-Flash reaches the
highest observed offloading accuracy of 1.00 with ReAct+few-shot+CoT, while
Llama3 obtains 0.92 with simpler CoT-based prompts but incurs substantially
higher latency. These results expose an accuracy--cost trade-off and show that
prompt components should be selected per model rather than accumulated
indiscriminately. Because each configuration contains only 25 cases and one
run, the values are treated as descriptive observations rather than stable
estimates of general model superiority.

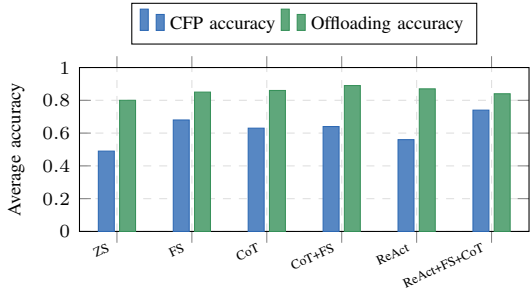
\begin{figure}[h!]
\centering
\begin{tikzpicture}
\begin{axis}[
    width=0.85\linewidth,
    height=3.75cm,
    ybar,
    bar width=6pt,
    ymin=0,
    ymax=1.0,
    ylabel={Average accuracy},
    symbolic x coords={ZS,FS,CoT,CoT+FS,ReAct,ReAct+FS+CoT},
    xtick=data,
    xticklabel style={rotate=25,anchor=east,font=\tiny},
    legend style={at={(0.5,1.14)},anchor=south,legend columns=2,font=\scriptsize},
    tick label style={font=\scriptsize},
    label style={font=\scriptsize},
    grid=major,
    grid style={dashed,gray!25},
]
\addplot+[fill=cfpblue!80,draw=cfpblue] coordinates {
(ZS,0.49)
(FS,0.68)
(CoT,0.63)
(CoT+FS,0.64)
(ReAct,0.56)
(ReAct+FS+CoT,0.74)
};

\addplot+[fill=offgreen!80,draw=offgreen] coordinates {
(ZS,0.80)
(FS,0.85)
(CoT,0.86)
(CoT+FS,0.89)
(ReAct,0.87)
(ReAct+FS+CoT,0.84)
};
\legend{CFP accuracy,Offloading accuracy}
\end{axis}
\end{tikzpicture}
\caption{Average impact of prompting strategies across the evaluated LLMs.}
\label{fig:scenario4b_prompting_average}
\end{figure}
\vspace{-5pts}
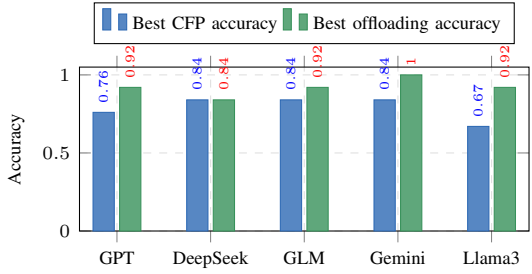
\begin{figure}[h!]
\centering
\begin{tikzpicture}
\begin{axis}[
    width=0.85\linewidth,
    height=3.75cm,
    ybar,
    bar width=8pt,
    ymin=0,
    ymax=1.05,
    ylabel={Accuracy},
    symbolic x coords={GPT,DeepSeek,GLM,Gemini,Llama3},
    xtick=data,
    legend style={at={(0.5,1.14)},anchor=south,legend columns=2,font=\scriptsize},
    tick label style={font=\scriptsize},
    label style={font=\scriptsize},
    nodes near coords,
    nodes near coords style={font=\tiny,rotate=90,anchor=west},
    grid=major,
    grid style={dashed,gray!25},
]
\addplot+[fill=cfpblue!80,draw=cfpblue] coordinates {
    (GPT,0.76) (DeepSeek,0.84) (GLM,0.84) (Gemini,0.84) (Llama3,0.67)
};
\addplot+[fill=offgreen!80,draw=offgreen] coordinates {
    (GPT,0.92) (DeepSeek,0.84) (GLM,0.92) (Gemini,1.00) (Llama3,0.92)
};
\legend{Best CFP accuracy,Best offloading accuracy}
\end{axis}
\end{tikzpicture}
\caption{Best observed CFP and offloading accuracy for each LLM.}
\label{fig:scenario4b_best_llm}
\end{figure}
\begin{figure}[h!]
\centering
\begin{tikzpicture}
\begin{axis}[
    width=0.82\linewidth,
    height=4cm,
    xlabel={Average token cost ($\times 10^3$)},
    ylabel={Average offloading accuracy},
    xmin=48,
    xmax=62,
    ymin=0.80,
    ymax=0.94,
    grid=major,
    grid style={dashed,gray!25},
    tick label style={font=\scriptsize},
    label style={font=\scriptsize},
    visualization depends on={value \thisrow{label} \as \pointlabel},
    nodes near coords={\pointlabel},
    nodes near coords style={font=\tiny,anchor=west},
]
\addplot[
    only marks,
    mark=*,
    mark size=2.8pt,
    color=costred
]
table[row sep=\\] {
x y label\\
52.79 0.80 ZS\\
55.02 0.85 FS\\
53.77 0.86 CoT\\
49.25 0.89 CoT+FS\\
60.38 0.87 ReAct\\
55.04 0.84 ReAct+FS+CoT\\
};
\end{axis}
\end{tikzpicture}
\caption{Average token-cost versus offloading-accuracy trade-off across prompting strategies.}
\label{fig:scenario4b_cost_tradeoff}
\end{figure}
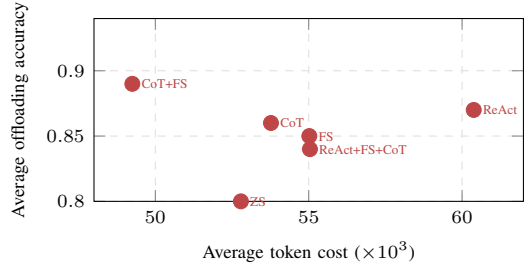
Figure~\ref{fig:scenario4b_prompting_average} shows that structured prompting consistently improves both CFP formulation and offloading accuracy. ReAct+Few-shot+CoT achieves the highest average CFP accuracy, whereas CoT+Few-shot delivers the highest average offloading accuracy, indicating that the best CFP formulation does not necessarily lead to the best end-to-end negotiation performance.
Figure~\ref{fig:scenario4b_best_llm} shows that Gemini-3-Flash achieves the highest offloading accuracy (1.00), while GLM-5.2 and GPT-OSS:20B also reach 0.92. In contrast, the best CFP accuracy varies across models, confirming that prompting effectiveness is LLM-dependent.
Figure~\ref{fig:scenario4b_cost_tradeoff} highlights the trade-off between negotiation performance and inference cost. \textit{CoT+Few-shot} provides the best balance, achieving the highest average offloading accuracy with the lowest token cost, whereas ReAct-based prompting incurs higher communication and inference costs without consistently improving offloading performance.


\begin{table}[h]
\centering
\caption{Experimental evidence for the proposed CNP extensions.}
\label{tab:cnp_evidence}
\scriptsize
\setlength{\tabcolsep}{1.5pt}
\begin{tabularx}{\columnwidth}{p{1.75cm}p{1.75cm}X}
\toprule
Extension & Evaluation contrast & Main observed evidence \\
\midrule
Multi-round revision & RB-SR vs. RB-MR & Drift violations: $0.53\!\rightarrow\!0.37$; overcommitment eliminated in all concurrent settings \\
Semantic contextual refinement & RB-MR vs. MAS-DecStream & Drift violations: $0.37\!\rightarrow\!0.03$; utility gain up to 22\%; conflict resolution $0.90\!\rightarrow\!0.95$ and $0.86\!\rightarrow\!0.91$ \\
Negotiation depth under fixed LLM & Large LLM, one vs. multiple rounds & Offloading accuracy: $0.71\!\rightarrow\!0.88$, with higher time and token cost \\
Deterministic safeguards & Multi-round pipelines & Zero overcommitment across the 5-, 10-, and 20-agent settings \\
\bottomrule
\end{tabularx}
\end{table}

Table~\ref{tab:cnp_evidence} consolidates how each experimental contrast relates to the protocol design. The table should not be read as a full component ablation: semantic context and LLM interpretation remain coupled in MAS-DecStream. It nevertheless shows that the largest, most consistent improvement comes from extending CNP with multiple validated refinement rounds, while LLM assistance provides an additional gain in context-dependent decisions.

\subsection{Cross-Scenario Discussion}
Three main findings emerge from the experiments. \textbf{First}, multi-round negotiation consistently provides the largest performance gain. It reduces scheduling violations under workload drift and eliminates resource overcommitment under concurrent requests, even without LLM assistance. This confirms that iterative proposal refinement improves decentralized decision-making compared with single-round negotiation~\cite{chen2025magicore,li2024sparsedebate}.
\textbf{Second}, the agentic-AI layer is most beneficial when decisions involve qualitative, ambiguous, or partially structured information. Rather than replacing deterministic scheduling, it improves contextual understanding and proposal refinement, consistent with recent LLM-based agentic systems~\cite{yu2026awto,wang2025maef,qiu2025blueprint}.
\textbf{Third}, deterministic verification remains essential. Agentic AI generates and refines negotiation proposals, whereas final scheduling decisions are validated against resource, QoS, and utility constraints before execution, preventing infeasible allocations~\cite{winston2026solver,qiu2025blueprint}.
These observations support an adaptive deployment strategy in which rule-based negotiation handles routine cases, while agentic AI is selectively invoked for uncertain, conflicting, or high-impact scheduling decisions. This limits inference latency, token consumption, and unnecessary negotiation overhead~\cite{ramirez2024optimising}.

\subsection{Threats to Validity}
\textbf{Construct validity.} CFP accuracy combines intent and contextual coverage and may not fully represent negotiation quality. We therefore report responder decisions, final-host accuracy, QoS violations, utility, and cost, and treat downstream scheduling outcomes as more important than message similarity.\\
\textbf{Internal validity.} Scenarios~1 and~2 compare complete pipelines. MAS-DecStream differs from RB-MR-CNP in both qualitative-context handling and LLM-assisted interpretation, so their difference is not a causal estimate of the LLM component. A component-matched rule-based context interpreter and LLM-without-context condition are required to isolate these effects.\\
\textbf{Conclusion validity.} The prompting experiment contains 25 cases and one run per model--prompt configuration. A single changed prediction therefore shifts accuracy by four percentage points, and selecting the best configuration among many comparisons may overstate performance. Repeated seeded runs, confidence intervals, and paired statistical tests are needed before using significance language.\\
\textbf{External validity.} The workloads are enriched from a production AI trace rather than executed in a physical stream-processing deployment. The largest configuration includes 20 representative agents and 12 concurrent requests, and the evaluated models and hosted endpoints may evolve. We release data-generation rules, prompts, model identifiers, and raw outputs to support replication, but larger topologies, network failures, state-transfer cost, and real operator migration remain future evaluation targets.

\section{Conclusion}\label{sec:conclusion}
This paper introduced LLM-MR-CNP, an extension of classical CNP with semantic CFP generation, progressive disclosure, iterative proposal revision, negotiation memory, and deterministic validation for decentralized stream-task offloading. Across the evaluated scenarios, extending the single proposal--award cycle to multiple validated rounds provides the largest and most consistent gain, reducing drift violations and eliminating overcommitment. LLM assistance adds smaller but useful improvements when refinement depends on ambiguous or qualitative runtime context and exhibits model-dependent accuracy--cost trade-offs. The incremental causal contribution of the LLM still requires component-matched ablations and repeated trials. Accordingly, this study provides an initial empirical assessment of a hybrid agentic scheduling architecture rather than proof of general superiority over established schedulers. Future work will evaluate selective LLM escalation, larger decentralized deployments, and stateful stream-operator migration under measured network and recovery costs.

\bibliographystyle{IEEEtran}
\bibliography{biblioConfAgenticAI}

\end{document}